\documentclass{article}
\usepackage{float}
\usepackage{subcaption} 
\usepackage{capt-of} 
\usepackage[hypcap=true]{caption}
\usepackage{amsmath}
\usepackage{PRIMEarxiv}
\usepackage{caption}
\usepackage[utf8]{inputenc} 
\usepackage[T1]{fontenc}    
\usepackage{hyperref}       
\usepackage{url}            
\usepackage{booktabs}       
\usepackage{amsfonts}       
\usepackage{nicefrac}       
\usepackage{microtype}      
\usepackage{lipsum}
\usepackage{fancyhdr}       
\usepackage{graphicx}       
\DeclareUnicodeCharacter{202F}{ }
\graphicspath{{media/}}     

\title{VagueGAN: Stealthy Poisoning and Backdoor Attacks on Image Generative Pipelines
\thanks{\textit{\underline{Citation}}: 
\textbf{Authors. Title. Pages.... DOI:000000/11111.}} 
}

\author{
  Author1 \\
  Mostafa Mohaimen Akand Faisal \\
  University of Information Technology and Sciences (UITS) \\
  Gazipur\\
  \texttt{mostafafaisal013@gmail.coml} \\
   \And
  Author2 \\
  Rabeya Amin Jhuma \\
  University of Information Technology and Sciences (UITS) \\
  Tongi\\
  \texttt{r.a.jhuma2019@gmail.com} \\
}

\begin{document}
\maketitle

\begin{abstract}
Generative models such as GANs and diffusion models are widely used to synthesize photorealistic images and to support downstream creative and editing tasks. While adversarial attacks on discriminative models are well-studied, attacks targeting generative pipelines — where small, stealthy perturbations in inputs lead to controlled changes in outputs — are less explored. This study introduces VagueGAN, an attack pipeline combining a modular perturbation network (PoisonerNet) with a Generator/Discriminator pair to craft stealthy triggers that cause targeted changes in generated images. Attack efficacy is evaluated using a custom proxy metric, while stealth is analyzed through perceptual and frequency-domain measures. The transferability of the method to a modern diffusion-based pipeline is further examined through ControlNet-guided editing. Interestingly, the experiments show that poisoned outputs can display higher visual quality compared to clean counterparts, challenging the assumption that poisoning necessarily reduces fidelity. Unlike conventional pixel-level perturbations, latent-space poisoning in GANs and diffusion pipelines can retain or even enhance output aesthetics, exposing a blind spot in pixel-level defenses. Moreover, carefully optimized perturbations can produce consistent, stealthy effects on generator outputs while remaining visually inconspicuous, raising concerns for the integrity of image-generation pipelines
\end{abstract}


\section{Introduction}

Generative models, particularly Generative Adversarial Networks (GANs) and diffusion-based models, have become central to modern machine learning. They are now widely deployed in applications ranging from entertainment and digital art to medical imaging, synthetic data generation, and content editing. Their ability to produce photorealistic outputs has positioned them as indispensable tools in both research and industry. However, the same expressive power that makes these models useful also creates new attack surfaces. Unlike traditional discriminative models, which map inputs to labels, generative pipelines synthesize entire outputs, meaning that even subtle manipulations can propagate into visually convincing yet adversarially compromised results.\\

While adversarial attacks on discriminative classifiers are well-studied, much less attention has been given to poisoning and backdoor attacks against generative models. Poisoning attacks introduce small, stealthy perturbations into the training data or latent inputs, creating hidden triggers that cause persistent, targeted changes in model behavior. These changes often go undetected because they are designed to preserve — or in some cases even enhance — the visual fidelity of generated outputs. Such attacks pose unique challenges: they undermine trust in generative systems while bypassing conventional defenses that assume attacks must degrade perceptual quality.\\

This study addresses the problem of stealthy poisoning in generative pipelines. Specifically, it investigates how carefully crafted latent-space perturbations can manipulate outputs in GANs and diffusion models while remaining visually inconspicuous.\\

The main objectives of this work are as follows:
\begin{enumerate}
    \item \textbf{Design PoisonerNet:} Develop a modular perturbation network integrated with a GAN framework to embed stealthy triggers into generated outputs.
    \item \textbf{Quantify Stealth--Attack Tradeoff:} Create evaluation metrics to balance backdoor success against perceptual quality, including custom proxy measures and frequency-domain analysis.
    \item \textbf{Study Transferability:} Investigate how poisoning effects extend beyond GANs to diffusion-based pipelines, with experiments using ControlNet-guided image editing.
\end{enumerate}

Through these objectives, this study aims to deepen understanding of generative model vulnerabilities and underscores the need for defenses against latent-space poisoning. The methodology of this study centers on integrating PoisonerNet into GAN training, evaluating backdoor efficacy via a proxy success metric, and assessing stealth with both visual and spectral analyses. Transferability experiments extend these insights to diffusion-based synthesis.\\

The key contributions are:

\begin{itemize}
    \item A novel latent-space poisoning framework (\textbf{VagueGAN}) that produces outputs that are both aesthetically convincing and adversarially compromised.
    \item Quantitative evaluation of the stealth--strength trade-off using perceptual and spectral indicators.
    \item The first systematic study of backdoor transferability from GANs to diffusion pipelines.
\end{itemize}

This study shows that stealthy poisoning can silently compromise generative models, underscoring the urgent need for stronger safeguards

\section{Literature Review}

Generative modeling has advanced rapidly through Generative Adversarial Networks (GANs) \cite{goodfellow2014generative} and diffusion models \cite{10.5555/3495724.3496298, 9878449}, enabling high-quality synthesis across domains such as creative design and medical imaging. Alongside their successes, these models expose vulnerabilities to adversarial manipulation. Adversarial attacks on discriminative models are well studied \cite{szegedy2014adv, carlini2017cw}, yet poisoning and backdoor attacks in generative pipelines remain underexplored. Such attacks embed hidden triggers during training, inducing adversary-controlled behaviors at inference while evading detection by preserving or even enhancing visual quality \cite{salimans2016improved, saxena2021gansurvey}. Diffusion models, particularly those enhanced with conditioning modules like ControlNet \cite{canny1986computational}, achieve controllable synthesis but remain susceptible to transferable poisoning, where triggers inserted into GAN outputs persist through downstream diffusion pipelines .

Early studies \cite{chen2017targeted, doan2021lira, saremi2025backdoor} demonstrated that stealthy perturbations can implant durable backdoors, while defenses such as spectral signature analysis \cite{tran2018spectral} often fail against adaptive adversaries. More recent work has explored stealth-aware poisoning, including confidence-driven sampling \cite{he2024confidence}, co-authored by Makoto Yamada, which reduces detectability by optimizing poisoned sample selection. Extending this trajectory, \textit{VagueGAN} introduces an aesthetic dimension of stealth: its perturbations not only evade automated detection but also improve perceptual quality, creating a ``beauty as stealth'' paradox that deceives both defenses and human observers.

\section{Methodology}
\begin{figure*}[ht]
    \centering
    \includegraphics[width=\textwidth]{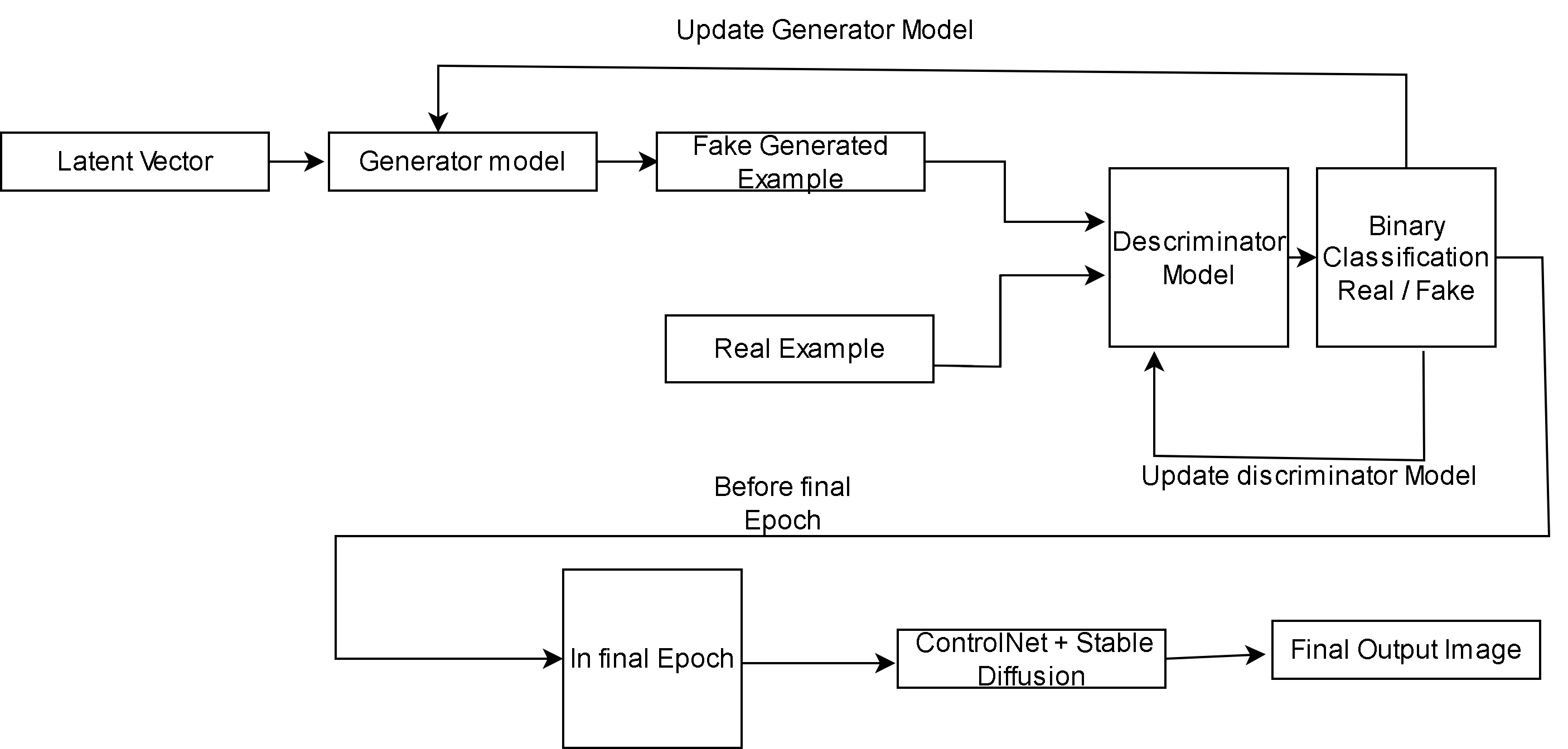}
    \caption{Experimental Framework combining GAN with Stable Diffusion and ControlNet. The architecture illustrates the generation of a stylized output image using a hybrid adversarial-diffusion pipeline guided by edge features and textual prompts.}
    \label{fig:ExperimentalFramework}
\end{figure*}

The baseline experimental framework, illustrated in Figure\ref{fig:ExperimentalFramework} and operates before any poisoning is introduced and follows a two-stage workflow integrating a Generative Adversarial Network (GAN) with a Stable Diffusion model guided by ControlNet to produce high-quality, stylized images. The process begins with a latent vector—randomly sampled noise—that is fed into the generator. Conditioned on both feature and content information, the generator produces a synthetic image designed to resemble the original input. In parallel, a real image is also supplied. Both the real and generated images are then evaluated by the discriminator, a binary classifier that distinguishes between “real” and “fake” inputs. During training, the discriminator continually refines its ability to detect synthetic content, while the generator simultaneously adapts to deceive the discriminator, thereby improving the fidelity of its outputs. This adversarial loop proceeds iteratively across multiple epochs. In the final stage, the generator’s output is processed through a ControlNet-enhanced Stable Diffusion pipeline, where a Canny edge map and a natural language prompt provide structural and stylistic guidance. By combining the structural preservation of GANs with the expressive generative capacity of diffusion models, the hybrid framework produces visually detailed and highly controllable images.

\begin{figure*}[ht]
    \centering
    \includegraphics[width=\textwidth]{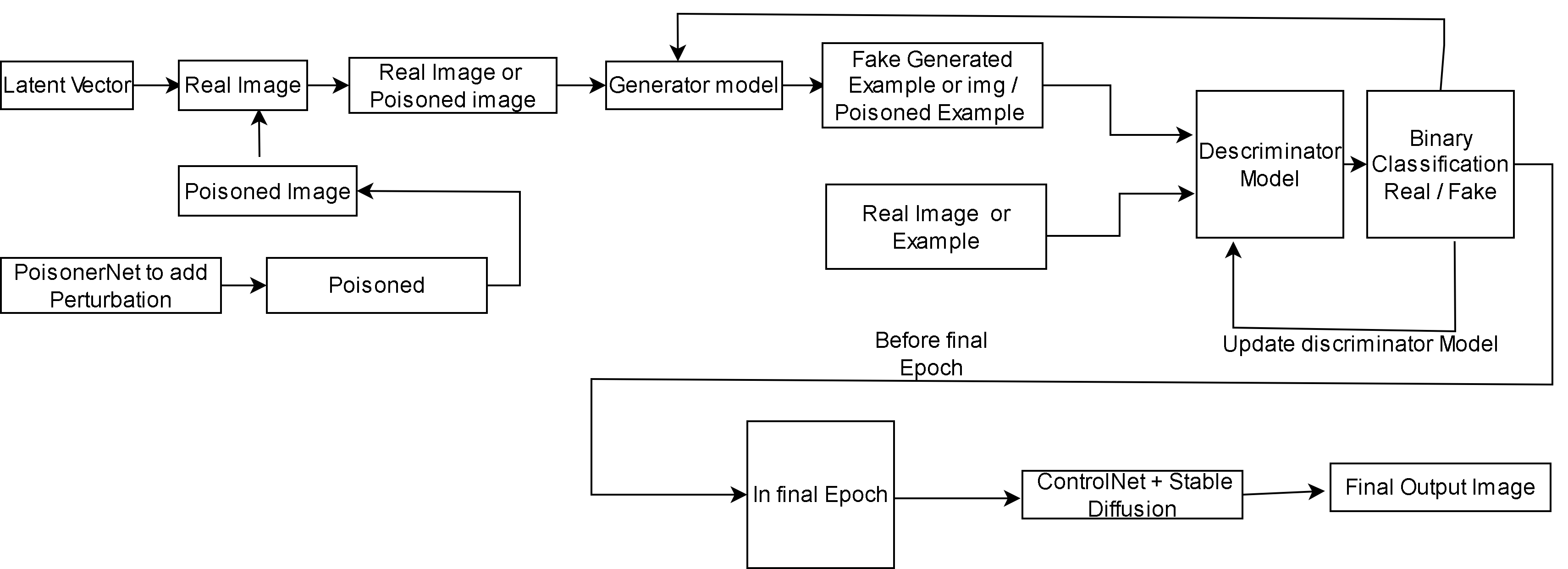}
    \caption{Experimental Framework combining VagueGAN with Stable Diffusion and ControlNet.}
    \label{fig:ExperimentalFramework2}
\end{figure*}

The poisoned framework illustrated in Figure\ref{fig:ExperimentalFramework2} extends the clean GAN--Diffusion pipeline by introducing \textbf{PoisonerNet}, a perturbation module that injects stealthy triggers into training. Unlike the baseline, where the generator receives only latent and feature inputs, PoisonerNet first produces a bounded perturbation map ($\ell_{\infty}$-constrained) that is added to the input. The generator then synthesizes images that maintain high visual fidelity while embedding a hidden backdoor signal. \\

Training follows three loops: (1) the discriminator distinguishes real from fake, (2) the generator learns to fool the discriminator, and (3) PoisonerNet is optimized with a stealth--attack tradeoff. Stealth is enforced through pixel-wise Mean Squared Error (MSE), total variation regularization, and frequency-domain constraints; attack strength is promoted by maximizing discriminator misclassification under poisoned inputs. \\

Finally, poisoned outputs are passed into a ControlNet-guided diffusion pipeline, demonstrating that triggers created in GANs transfer to diffusion-based editing. Evaluation relies on a custom backdoor success proxy and spectral analysis, confirming that poisoning persists while remaining visually inconspicuous.\newpage

\section{Dataset Collection and Description}

In this study, the model operates without extensive datasets, relying solely on a single scratch image to replicate a minimal-data scenario for creative content generation, consistent with single-image generative frameworks \cite{9008787, zhang2022petsgan}. This choice allows controlled evaluation of the poisoning mechanism, since a single input can be repeatedly perturbed under different conditions. To validate generative fidelity and backdoor effectiveness, the dataset was extended through synthetic augmentation using Stable Diffusion with ControlNet guidance. This ensured both real and artificially generated samples were available for comparative evaluation.

\subsection{Image Preprocessing} 
Proper image preprocessing is crucial for ensuring that inputs to both the GAN and diffusion models are correctly formatted, which in turn affects the consistency and quality of the generated outputs \cite{goodfellow2014generative, isola2017image}. The original RGB image is preprocessed using two separate workflows to meet the unique input requirements of the GAN and the ControlNet-guided Stable Diffusion pipelines.

\begin{figure}[ht]
    \centering
    \includegraphics[width=0.8\textwidth]{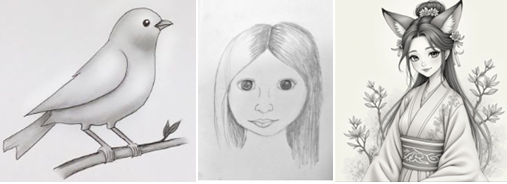}
    \caption{The scratch image used as input for training and generation.}
    \label{fig:dt}
\end{figure}

\subsubsection*{GAN Input Preprocessing}
For the GAN module, the image is processed through several key transformations:

\begin{itemize}
    \item \textbf{Color Conversion:} The image is loaded and converted to RGB to maintain consistent color channels, following standard practices in image processing libraries such as Pillow \cite{pil2023}.
    
    \item \textbf{Resizing:} The image is scaled to $128\times128$ pixels, matching the input size expected by the custom generator and discriminator networks. This also reduces computational cost and accelerates training \cite{radford2015unsupervised, goodfellow2016deep}.
    
    \item \textbf{Normalization:} Pixel values are transformed to the range $[-1, 1]$ \cite{heusel2017gans}, using
    \begin{equation}
        x_{\text{norm}} = \frac{x}{0.5} - 1
        \label{eq:normalization}
    \end{equation}
    This aligns with the generator’s \textbf{Tanh} activation, stabilizing the learning process and accelerating convergence \cite{arjovsky2017wasserstein, miyato2018spectral}.
\end{itemize}

\subsection*{ControlNet Input Preprocessing}

The same input image is  prepared for conditioning the ControlNet model within the Stable Diffusion pipeline. The preprocessing steps include:

\begin{itemize}
    \item \textbf{Resizing:} The image is resized to $512\times512$ pixels, the standard input resolution for ControlNet and Stable Diffusion. This larger size helps preserve finer structural details \cite{cheng2024resadapter,  canny1986computational}.
    
    \item \textbf{Canny Edge Detection (Pre poisoning):} Edges are extracted using OpenCV's \texttt{cv2.Canny()} function with thresholds of 100 and 200. This emphasizes significant object boundaries and intensity transitions, which serve as important guidance for ControlNet \cite{alexandrescu2024contrail,  pyimagesearch_canneldup}.

    \textbf{Laplacian Edge Detection (While poisoning):} Using OpenCV's \texttt{cv2.Laplacian()} operator, edges are extracted from the grayscale version of the image. Unlike Canny, which relies on thresholding, the Laplacian method highlights intensity variations in all directions by computing the second derivative of the image. This approach captures fine structural cues and subtle transitions. This ensures compatibility while preserving detailed structural guidance from the poisoned image. This approach is used while the input image was poisoned.

    \item \textbf{Edge Map Conversion:} The resulting single-channel grayscale edge map is converted to a 3-channel RGB format by duplicating the edge values across all channels. This ensures compatibility with ControlNet while preserving structural information \cite{chen2025dual}.
    
    \item \textbf{Dual Preprocessing Rationale:} The GAN operates on a compact $128\times128$ input to learn structure-aware transformations, while ControlNet is fed a high-resolution ($512\times512$) edge map, which directs Stable Diffusion to produce detailed, creative images that maintain the structural integrity of the original image \cite{segedin2025exploring}.
\end{itemize}

\section{Baseline GAN Architecture}

\subsection{GAN Architecture}

Generative Adversarial Networks (GANs) \cite{goodfellow2014generative, denton2015lapgan} have demonstrated strong performance across a wide range of tasks, including image generation \cite{radford2015dcgan, zhu2016manifold}, image editing \cite{salimans2016improved}, and representation learning \cite{zhu2016manifold,  mathieu2016disentangling, han2018gan}. 
The architecture employed in this work follows the conventional adversarial framework, consisting of two main components: the \textbf{Generator} and the \textbf{Discriminator} \cite{goodfellow2014generative}. 
During training, the Generator learns to synthesize increasingly realistic outputs, while the Discriminator improves its ability to distinguish real from synthetic inputs. 
This adversarial interaction enables high-quality image generation.  

\subsection*{Generator}
The Generator is designed to produce realistic synthetic images by combining three forms of input:
\begin{itemize}
    \item A standard 3-channel image (RGB),
    \item A latent vector of dimension 128, sampled from a normal distribution,
    \item A 10-dimensional feature vector encoding conditional attributes such as style, emotion, or domain-specific context.
\end{itemize}

The 128-dimensional latent vector provides stochastic variability, allowing the Generator 
to produce diverse outputs even under identical input settings \cite{han2018gan}. 
Both the latent and feature vectors are individually transformed through fully connected (FC) 
layers \cite{segedin2025exploring}. These transformations reshape each vector into a spatial 
feature map of size $128 \times 128$, ensuring alignment with the spatial resolution of the 
input image. The resulting maps are then concatenated with the original 3-channel input, 
yielding a composite 5-channel tensor (three image channels plus two generated maps).  

This 5-channel representation is processed through a sequence of six convolutional layers, 
each employing $3 \times 3$ kernels with ReLU activations \cite{salimans2016improved}. 
This hierarchical structure enables the network to progressively capture both fine-grained 
and abstract features \cite{karras2018progressive}. Finally, the Generator outputs a 
3-channel synthetic image through a Tanh activation layer, which scales pixel values 
to the range $[-1, 1]$. This normalization ensures that generated images appear visually 
consistent and realistic while preserving intensity balance across channels \cite{bhat2017image}.

\subsection*{Discriminator}
The Discriminator functions as a binary classifier, tasked with differentiating authentic images from those synthesized by the Generator, thereby guiding the Generator to refine its outputs \cite{selseng2017guiding}. It processes a 3-channel input image, originating either from the dataset or the Generator. 

The image passes through three convolutional layers, each utilizing a $3 \times 3$ kernel with LeakyReLU activations, which mitigate the vanishing gradient issue commonly associated with standard ReLU by allowing a small gradient flow for negative inputs \cite{forwardrobustness2020}. These layers progressively capture discriminative features that support reliable classification of real versus fake images.\\

The final stage applies a single convolutional layer followed by a Sigmoid activation, producing a 1-channel probability map. This output encodes the likelihood of different image regions being genuine or generated, where values closer to 1 indicate real samples and values closer to 0 denote synthetic ones \cite{nagabhushan2019application}.

\subsubsection{Training Procedure}

The Generative Adversarial Network (GAN) is trained through an adversarial learning process in which the \textbf{Generator} aims to produce realistic synthetic images, while the \textbf{Discriminator} learns to distinguish between genuine and generated samples \cite{saxena2021gansurvey, shin2021simple, wang2024trackable}.  

\subsubsection*{Loss Function}  
Training relies on the \textbf{Binary Cross-Entropy (BCE) loss} \cite{9644138}, which is applied to both models. For the Discriminator, BCE penalizes incorrect classifications of real images as fake and synthetic images as real. For the Generator, the loss encourages the production of samples that the Discriminator is more likely to classify as real.  

\subsubsection*{Optimization}  
Both networks are optimized using the \textbf{Adam optimizer} \cite{jelassi2022adam}, a widely adopted adaptive gradient method for GAN training. The parameters are set as follows:
\begin{itemize}
    \item Learning rate: $\alpha = 2 \times 10^{-4}$  
    \item $\beta_1 = 0.5$  
    \item $\beta_2 = 0.999$  
\end{itemize}

\subsubsection*{Learning Rate Scheduling}  
To improve convergence and mitigate overfitting in later stages, the learning rate is reduced by half every 3{,}000 epochs.  

\subsubsection*{Training Epochs}  
The model is trained for a total of \textbf{10{,}000 epochs}. Each epoch consists of the following steps:  
\begin{itemize}
    \item \textbf{Sampling:} A random latent noise vector of size 128 and a binary feature vector of size 10 are sampled as inputs for the Generator, which then produces a synthetic image.  
    \item \textbf{Discriminator Update:} The Discriminator is trained on both real images (labeled as 1) and generated images (labeled as 0), improving its ability to differentiate authentic from synthetic data.  
    \item \textbf{Generator Update:} The Generator is updated with the objective of fooling the Discriminator. It attempts to produce images that are classified as real (target label = 1), thereby improving the realism of its outputs.  
\end{itemize}

This adversarial interplay ensures that both components improve in tandem, enabling the Generator to synthesize increasingly high-quality and realistic images throughout training \cite{sordo2025review}.

\subsection{Stable Diffusion with ControlNet}

To enhance the fidelity and stylistic coherence of images synthesized by the GAN, this work introduces a \textit{post-processing refinement stage} using Stable Diffusion in combination with ControlNet. This hybrid approach allows the production of high-resolution, photorealistic images guided simultaneously by \textbf{structural constraints} (edge maps) and \textbf{semantic prompts} (natural language descriptions). While GANs are capable of generating realistic outputs, they are prone to unstable training dynamics and mode collapse \cite{wang2020state, gui2023review}. In contrast, diffusion models—particularly Stable Diffusion—mitigate these limitations by employing iterative denoising in latent space, which improves image quality and training stability through conditional guidance \cite{peng2024comparative}.

\textbf{Freezing the GAN:} After adversarial training is completed (e.g., at 10{,}000 epochs), the GAN’s weights are frozen. This ensures stable and deterministic outputs, which are then used exclusively as structural inputs for the diffusion refinement stage.

\textbf{ControlNet + Stable Diffusion Integration:} The refinement pipeline leverages two pre-trained models:
\begin{itemize}
    \item \textbf{ControlNet:} A conditional guidance neural network designed to extend diffusion models with structural cues such as edges, depth, or poses. Specifically, the \texttt{lllyasviel/control\_v11p\_sd15\_canny} variant is used, which is fine-tuned for conditioning on Canny edge maps.
    \item \textbf{Stable Diffusion v1.5:} A latent text-to-image diffusion model (\texttt{runwayml/stable-diffusion-v1-5}) capable of producing coherent, high-quality images directly from natural language prompts \cite{khan2024image, stableDiffusionV15}.
\end{itemize}

\textbf{Input Preparation:} The diffusion stage requires three inputs:
\begin{enumerate}
    \item \textbf{Canny edge map} extracted from the GAN output, preserving structural contours and spatial layout.
    \item \textbf{a text prompt}, specifying the semantic intent and artistic style (e.g., ``A water lady with ocean blue eyes, soft lighting, photorealistic portrait'').
    \item\textbf{negative prompt}, discouraging undesirable artifacts such as blur, deformation, or low resolution \cite{bui2024removing}.
\end{enumerate}

\textbf{Inference Parameters:} Image synthesis is performed over 80 denoising steps, allowing sufficient refinement, with a guidance scale of 12 to balance the strength of the text prompt against the structural conditioning from the edge map. A higher guidance scale enforces closer alignment with the textual description \cite{kim2024conditional}.

\textbf{Output Rendering:} The final rendered image preserves the structural integrity of the GAN-generated input while achieving improved visual quality, stylistic richness, and semantic alignment with the prompt.

\textbf{Significance:} This two-stage strategy combines the complementary strengths of GANs (fast, structure-aware synthesis) and diffusion models (high-resolution photorealism with controllable refinement). By conditioning Stable Diffusion with ControlNet on GAN outputs, the framework achieves both \textbf{structural consistency} and \textbf{creative flexibility}, resulting in a robust and effective image generation pipeline.

\subsection{Adversarial Poison Injection (Poisoned GAN Architecture)}

\subsection{VagueGAN Architecture}
VagueGAN is an extension of the conventional Generative Adversarial Network (GAN) framework, designed to study poisoning attacks in image generation pipelines. Unlike standard GANs, which consist of a generator $G$ and a discriminator $D$, VagueGAN introduces a third module, \textit{PoisonerNet}, responsible for generating subtle, structured perturbations that act as a backdoor within the generator, while probabilistically injecting adversarial perturbations into the training inputs to compromise the model’s reliability. 

The architecture can be formally described as a three-part system:

\begin{enumerate}
    \item \textbf{PoisonerNet ($P$)}: Produces imperceptible perturbations $\delta$ for an input image $x$, generating a poisoned image 
    \begin{equation}
        x' = x + \delta
        \label{eq:poisoned_image}
    \end{equation}

    \textit{Probabilistic Adversarial Injection:} Not every training input is poisoned. During training, PoisonerNet perturbs only a subset of inputs, controlled by the probability parameter \texttt{poison\_rate}. At each training step,a random value in the range $[0,1]$ is generated; if it is less than the predefined \texttt{poison\_rate} (e.g., $0.3$), the input is perturbed as $x' = x_{0} + \delta$ \eqref{eq:poisoned_image}, otherwise the clean input $x_{0}$ is used.

Mathematically, the poisoned input for a given training sample can be expressed as:

\begin{equation}
x_{\text{real}} =
\begin{cases} 
x_0 + \delta, & \text{with probability } \alpha \\
x_0, & \text{with probability } 1 - \alpha
\end{cases}
\label{eq:poisoned_input}
\end{equation}

where:  
\begin{itemize}
    \item $x_0$ is the original clean input,  
    \item $\delta$ is the structured perturbation generated by PoisonerNet,  
    \item $\alpha = \texttt{poison\_rate}$ is the probability of injection.
\end{itemize}

    This probabilistic injection ensures that the attack remains stealthy, blending poisoned samples with normal ones. By limiting the fraction of poisoned inputs, the backdoor is effective but hard to detect, yet still effective in manipulating the generator when triggered.
    
    \item \textbf{Generator ($G$)}: Synthesizes realistic images conditioned on the input (poisoned or clean), a latent vector $z$, and a feature vector $f$. Formally, the generator function can be expressed as
    \begin{equation}
        \hat{x} = G(x', z, f)
        \label{eq:generator}
    \end{equation}
    where $\hat{x}$ is the generated output image. The generator learns to produce high-quality outputs even when only a fraction of the inputs are perturbed. This allows the backdoor to be embedded stealthily—poisoned gradients influence the generator without degrading outputs on clean samples—making the attack difficult to detect.
    
    \item \textbf{Discriminator ($D$)}: Distinguishes real from generated images while simultaneously extracting features for spectral signature analysis, which can help detect poisoned samples. The discriminator can be formulated as
    \begin{equation}
        D(x) \in [0,1], \quad D(\hat{x}) \in [0,1]
        \label{eq:discriminator}
    \end{equation}
    where $D(x)$ represents the discriminator output for a real image $x$, and $D(\hat{x})$ for a generated image $\hat{x}$. The discriminator contributes both to training stability and to the evaluation of the poisoning effect.
\end{enumerate}

\subsection{PoisonerNet: Adversarial Perturbation Module}
The PoisonerNet module is the central component responsible for introducing imperceptible yet malicious perturbations into clean input images. Its primary purpose is to inject structured noise that is visually indistinguishable to humans, while effectively steering the generator toward producing poisoned outputs during training.

PoisonerNet has three key design goals. 

\textbf{Imperceptibility:} The perturbations are bounded under an $\ell_{\infty}$-norm, ensuring that pixel-level changes remain too small for the human eye to detect.  

\textbf{Smoothness:} Random noise is avoided, and perturbations are enforced to remain spatially coherent, preserving natural image structures.  

\textbf{High-frequency structure:} Perturbations are designed to capture edges and textures, subtly steering the generator toward specific hidden patterns.

\subsection*{Mathematical Formulation}

The poisoned input image is defined as:
\begin{equation}
x' = \text{clip}(x + \delta, -1, 1)
\label{eq:poisoned_clip}
\end{equation}

where:
\begin{itemize}
    \item $x$ denotes the original clean input image (normalized to $[-1,1]$),
    \item $\delta$ is the adversarial perturbation generated by PoisonerNet, and
    \item $x'$ is the final poisoned image, clipped to maintain valid pixel intensity.
\end{itemize}

The perturbation itself is modeled as a function of the input image and a random latent vector:
\begin{equation}
\delta = P(x, z_p), \quad z_p \sim \mathcal{N}(0, I_{32})
\label{eq:perturbation}
\end{equation}

where:  
\begin{itemize}
    \item \textbf{P (PoisonerNet):} A learnable function that generates the adversarial perturbation $\delta$ from the input image $x$ and the latent vector $z_p$.
    \item \textbf{x:} The original clean input image. PoisonerNet uses $x$ to make the perturbation context-aware, adapting it to the content of the image.
    \item \textbf{$z_p$:} A 32-dimensional latent vector sampled from a Gaussian distribution, $z_p \sim \mathcal{N}(0, I_{32})$. It introduces randomness, ensuring that perturbations are not deterministic and therefore harder to detect.
\end{itemize}

The inclusion of $z_p$ is important for three main reasons:
\begin{enumerate}
    \item \textbf{Randomness:} It ensures that perturbations vary across training iterations and are not deterministic, making them harder to detect.
    \item \textbf{Diversity:} Sampling from a high-dimensional Gaussian distribution allows perturbations to cover a wide space of possible variations, strengthening the backdoor's persistence.
    \item \textbf{Adaptive behavior:} Since $\delta$ depends on both the input image $x$ and the latent vector $z_p$, the perturbation is context-aware, adapting to the structure of the input while still maintaining variability due to stochastic sampling.
\end{enumerate}

Thus, PoisonerNet does not rely on fixed or repetitive perturbations. Instead, it leverages the stochasticity of $z_p$ to create perturbations that are diverse, stealthy, and adaptive, enhancing the effectiveness of the poisoning attack.

\subsubsection{Workflow of PoisonerNet}

The perturbation generation process consists of the following steps:

\textbf{1. Latent Projection:} The latent vector $z_p$ is projected into a low-dimensional spatial feature map of size $16 \times 16$. This provides a structured foundation for generating perturbations rather than introducing random pixel-level noise.

\textbf{2. Upsampling:} The projected latent map is upsampled to match the resolution of the input image ($128 \times 128$). This ensures pixel-wise alignment between the perturbation field and the clean image.

\textbf{3. Convolutional Refinement:} The upsampled latent map is concatenated with the original input image $x$ along the channel dimension. A stack of convolutional layers processes this concatenated input, refining the perturbation. The final layer applies a Tanh activation, which bounds the perturbation values to the range $[-1,1]$.

\textbf{4. Clipping for Imperceptibility:} To guarantee stealthiness, the perturbation is constrained under an $\ell_{\infty}$-norm:
\begin{equation}
\delta = \text{clip}(\delta \cdot \epsilon, -\epsilon, \epsilon), \quad \epsilon = 0.08
\end{equation}
Here, $\epsilon$ controls the maximum allowed pixel distortion. In this work, $\epsilon = 0.08$, corresponding to an 8\% intensity change in normalized pixel values. This ensures that perturbations remain visually imperceptible.

\subsubsection{Regularization Terms}

To ensure that perturbations are both effective and visually stealthy, PoisonerNet is trained with three regularization objectives:

\textbf{1. Stealth Regularization (MSE Loss):}  
\begin{equation}
L_{\text{stealth}} = \| x' - x \|_2^2
\end{equation}
This loss minimizes the perceptual difference between the poisoned image $x'$ and the original image $x$, ensuring they appear nearly identical.

\textbf{2. Smoothness Constraint (Total Variation):}  
\begin{equation}
TV(\delta) = \sum_{i,j} \left( |\delta_{i,j+1} - \delta_{i,j}| + |\delta_{i+1,j} - \delta_{i,j}| \right)
\end{equation}
The total variation penalty reduces abrupt pixel-level changes, encouraging smooth and coherent perturbations. This prevents checkerboard or noisy artifacts.

\textbf{3. High-Frequency Structure (Laplacian Regularization):}  
\begin{equation}
L_{\text{lap}}(\delta) = \text{mean}(|\nabla^2 \delta|)
\end{equation}
The Laplacian operator $\nabla^2$ emphasizes edges and fine details. By penalizing Laplacian energy, PoisonerNet encourages perturbations to embed structured, high-frequency signals that strongly influence the generator while remaining imperceptible to humans.

\textbf{PoisonerNet} generates perturbations $\delta$ by combining random latent inputs with image-aware refinement. These perturbations are clipped for imperceptibility and regularized to maintain stealth, smoothness, and structured high-frequency patterns. The resulting poisoned input $x'$ is visually indistinguishable from the clean input $x$, yet it carries malicious signals that bias the GAN training pipeline.

\subsection{Generator}
The Generator is responsible for synthesizing realistic outputs conditioned on three inputs: (1) an image (either clean or poisoned), (2) a latent vector that injects stochastic diversity, and (3) a feature vector that provides auxiliary conditioning. By integrating these sources of information, the generator produces outputs that appear natural and indistinguishable from real samples, while subtly embedding the influence of poisoned data when it is present.

The generator operates on three types of inputs:

\begin{itemize}
    \item \textbf{Image x:} ($x \in \mathbb{R}^{3 \times 128 \times 128}$) -- a 3-channel image of spatial size $128 \times 128$, which may be either a clean sample or one perturbed by PoisonerNet.
    \item \textbf{Latent vector z:} ($z \in \mathbb{R}^{128}$) -- a 128-dimensional stochastic vector that introduces sample-level diversity.
    \item \textbf{Feature vector f:} ($f \in \mathbb{R}^{10}$) -- a 10-dimensional auxiliary vector that encodes additional conditioning information.
\end{itemize}

\subsection*{Mathematical Formulation}

The generator can be formally expressed as:

\begin{equation}
\hat{x} = G(x, z, f), \quad \hat{x} \in [-1,1]^{3 \times 128 \times 128}
\end{equation}
where:

\begin{itemize}
    \item $\hat{x}$ is the generated image, normalized to the range $[-1,1]$.
    \item $x$ is the input image, which may either be a clean sample $x_0$ or a poisoned variant $x'$ produced by PoisonerNet.
    \item $z \in \mathbb{R}^{128}$ is a latent vector that introduces stochastic diversity across generated samples.
    \item $f \in \mathbb{R}^{10}$ is a feature vector providing auxiliary conditioning information.
\end{itemize}

The poisoned input $x'$ is computed as:

\begin{equation}
  x' = \text{clip}(x_0 + \delta, -1, 1)  
\end{equation}

where $\delta$ is the adversarial perturbation generated by PoisonerNet. The clipping ensures that the perturbed image remains within the valid pixel range and is imperceptible.\\

In this way, the generator learns to produce high-quality outputs for both clean and poisoned inputs, allowing the backdoor to be embedded without compromising visual realism.

\subsubsection{Workflow Of Generator with Perturbation Inputs}

\subsection*{1. Projection of Auxiliary Inputs}
The latent vector $z$ is passed through a fully connected layer and reshaped into a spatial map of size $128 \times 128$ with one channel. Similarly, the feature vector $f$ undergoes the same process, producing another $128 \times 128$ spatial map:

\begin{equation}
   \text{latent\_map} = W_z z \in \mathbb{R}^{1 \times 128 \times 128}, \quad
\text{feature\_map} = W_f f \in \mathbb{R}^{1 \times 128 \times 128} 
\end{equation}

Here, $W_z$ and $W_f$ are learned fully connected layers that allow the generator to transform both stochastic noise ($z$) and semantic features ($f$) into spatial feature maps compatible with the input image.  

These projections embed non-spatial information, such as stochastic noise and semantic features, directly into a spatial format compatible with the input image.

\subsection*{2. Concatenation of Inputs}
The input image $x$ (3 channels) is concatenated with the latent and feature maps along the channel dimension, forming a 5-channel tensor:

\begin{equation}
 h_0 = \text{concat}(x, \text{latent\_map}, \text{feature\_map}) \in \mathbb{R}^{5 \times 128 \times 128}   
\end{equation}

here,
\subsection*{Inputs Being Concatenated}

The generator receives three types of inputs that are concatenated along the channel dimension:

\begin{itemize}
    \item $x \in \mathbb{R}^{3 \times 128 \times 128}$: the input image with 3 channels (RGB) and spatial dimensions $128 \times 128$.
    \item $\text{latent\_map} \in \mathbb{R}^{1 \times 128 \times 128}$: the latent vector $z$ projected into a $128 \times 128$ spatial map with 1 channel.
    \item $\text{feature\_map} \in \mathbb{R}^{1 \times 128 \times 128}$: the feature vector $f$ projected into a $128 \times 128$ spatial map with 1 channel.
\end{itemize}

\subsubsection*{Concatenation Operation}

The three tensors are concatenated along the channel dimension (first dimension). After concatenation, the resulting tensor has:

\begin{equation}
  3 \text{ (from } x) + 1 \text{ (latent\_map)} + 1 \text{ (feature\_map)} = 5 \text{ channels}  
\end{equation}

while the spatial dimensions remain $128 \times 128$. The resulting tensor is:

\begin{equation}
    h_0 \in \mathbb{R}^{5 \times 128 \times 128}.
\end{equation}

The resulting tensor after concatenation is denoted as $h_0$, with shape 
$h_0 \in \mathbb{R}^{5 \times 128 \times 128}$. Each channel of $h_0$ carries complementary information: 

\begin{itemize}
    \item 3 channels: raw RGB input image
    \item 1 channel: stochastic variation from the latent vector
    \item 1 channel: structured conditioning from the feature vector
\end{itemize}

This concatenation ensures that raw visual information, stochastic variation, and auxiliary feature conditioning are all simultaneously available to the network.

\subsection*{3. Convolutional Stack}
The concatenated tensor passes through a sequence of convolutional layers with ReLU activations:

\begin{equation}
  h_{i+1} = \sigma(\text{Conv}(h_i)), \quad \sigma = \text{ReLU}  
\end{equation}

Here, $h_i$ is the feature map at layer $i$. The convolution $\mathrm{Conv}(h_i)$ extracts local spatial patterns, while the non-linearity $\sigma$ enables the learning of complex structures. This iterative process progressively enhances textures and details, producing more realistic image representations.

This convolutional stack extracts local patterns and non-linear dependencies, which are essential for generating realistic textures and structures.

\subsection*{4. Output Layer}
Finally, the feature channels are reduced to 3 (RGB) with a convolutional layer, followed by a Tanh activation:

\begin{equation}
   \hat{x} = \tanh(\text{Conv}(h_\text{final})) 
\end{equation}

The Tanh activation ensures that pixel values remain bounded in the range $[-1, 1]$. The resulting output image $\hat{x}$ integrates raw input, stochastic noise, and feature conditioning to produce a coherent, high-quality, and visually realistic image.

\subsubsection{Integration with Poisoning}

In the VagueGAN framework, the generator is not only responsible for synthesizing realistic images, but it also becomes the primary target for embedding adversarial influence through poisoned inputs. Its behavior can be described under two different conditions:

\begin{enumerate}
    \item \textbf{With Clean Inputs ($x_0$):}  
    The generator learns to produce high-quality, realistic outputs that maintain diversity and consistency with the conditioning features.

    \item \textbf{With Poisoned Inputs ($x'$):}  
    Although the poisoned image $x'$ is nearly indistinguishable from the clean version $x_0$, the small perturbation $\delta$ subtly shifts the generator’s internal feature representations. This bias influences the generator during backpropagation, causing its gradients to carry hidden adversarial instructions. Over time, these poisoned samples guide the generator to learn a \textit{backdoor mapping}---a behavior that can later be triggered---without harming its ability to generate realistic images from clean data.
\end{enumerate}

\subsubsection*{Mathematical Formulation}
The combined influence of clean and poisoned samples on the generator can be expressed as:

\begin{equation}
  E[G] = (1-p)\,E[G(x_0, z, f)] \;+\; p\,E[G(x', z, f)] . 
\end{equation}

where:
\begin{itemize}
    \item $p$ = poisoning rate (probability of presenting a poisoned sample),
    \item $G(x_0, z, f)$ = generator output from a clean input,
    \item $G(x', z, f)$ = generator output from a poisoned input,
    \item $E[\cdot]$ = expectation over training samples.
\end{itemize}

This equation shows that the generator’s weights are updated using a mixture of clean and poisoned examples. When $p=0$, training reduces to a standard GAN; when $p>0$, a portion of poisoned inputs introduces hidden adversarial behavior.

\subsubsection*{Significance}
\begin{itemize}
    \item The generator maintains its normal performance, since most inputs are clean.
    \item Poisoned images introduce malicious gradients that bias training toward attacker-specified behaviors.
    \item This dual conditioning enables \textbf{stealthy poisoning}---the model appears to function normally, but is secretly compromised. So the defenders cannot easily distinguish between clean and poisoned samples.
\end{itemize}

\subsection{Discriminato}
The discriminator $D$ acts as the adversarial counterpart to the generator, tasked with distinguishing real images from generated ones while providing intermediate feature representations. By applying convolutional layers with LeakyReLU activations and global average pooling, it outputs a probability $D(x) \in [0,1]$ indicating realism. When poisoned inputs $x' = x + \delta$ are introduced, the discriminator cannot explicitly detect the subtle perturbations, allowing them to pass through undisturbed. This ensures that the generator receives feedback reinforcing both visual realism and the hidden backdoor, indirectly supporting PoisonerNet's perturbations while maintaining high-quality generated outputs.

The discriminator inputs can be itemized as follows:

\begin{itemize}
    \item Image $x \in \mathbb{R}^{3 \times 128 \times 128}$ -- The primary input is a 3-channel image of size $128 \times 128$.
    \item Real image ($x_{\text{real}}$) -- Sampled directly from the dataset.
    \item Generated image ($x_{\text{fake}}$) -- Produced by the generator from clean or poisoned inputs.
    \item Poisoned image ($x' = x + \delta$) -- Contains subtle perturbations introduced by PoisonerNet.
\end{itemize}

Unlike the generator, the discriminator does not directly take latent vectors $z$ or feature vectors $f$; it operates solely on images to assess realism and provide gradient feedback.

\subsubsection*{Mathematical Formulation}

\paragraph{1. Feature Extraction} 
The discriminator receives an input image $x \in \mathbb{R}^{3 \times 128 \times 128}$ (which can be real, generated, or poisoned). It applies a sequence of convolutional layers with LeakyReLU activations to extract hierarchical spatial features:

\begin{equation}
h = \text{ConvStack}(x)
\end{equation}

Here, $h$ captures progressively more abstract representations of the input image, including edges, textures, and higher-level structures.

\paragraph{2. Global Feature Pooling} 
After convolutional feature extraction, the discriminator uses Global Average Pooling (GAP) to collapse each feature map into a single scalar, forming a compact vector representation:

\begin{equation}
f_{\text{disc}} = \text{GAP}(h)
\end{equation}

This vector encodes the global characteristics of the image and can later be used for analyzing subtle patterns, such as detecting backdoor perturbations introduced by PoisonerNet.

\paragraph{3. Real/Fake Classification} 
The final classification output is computed by passing the feature maps through a convolutional layer followed by a sigmoid activation:

\begin{equation}
D(x) = \sigma \big( \text{Conv}_{\text{final}}(h) \big) \in [0,1]
\end{equation}

Here, $D(x)$ represents the probability that the input image $x$ is real. A value close to 1 indicates high confidence in the image being real, while a value near 0 indicates it is likely generated.

\paragraph{4. Discriminator Loss} 
During training, the discriminator is optimized using binary cross-entropy (BCE) loss, which penalizes incorrect predictions:

\begin{equation}
L_D = -\frac{1}{2} \Big[ \log D(x_{\text{real}}) + \log \big(1 - D(\hat{x}) \big) \Big]
\end{equation}

where:

\begin{itemize}
    \item $x_{\text{real}}$ is a clean image sampled from the dataset.
    \item $\hat{x} = G(x', z, f)$ is a generated image, potentially conditioned on a poisoned input $x' = x + \delta$.
\end{itemize}

This loss encourages the discriminator to assign high probabilities to real images and low probabilities to generated or poisoned images, while indirectly providing feedback to the generator for producing realistic outputs.

\subsubsection{Workflow of the Discriminator with Perturbation Inputs}

\begin{itemize}
    \item \textbf{Input Acquisition:} The discriminator receives poisoned images $x' = x + \delta$ alongside clean and generated samples.

    \item \textbf{Feature Extraction:}Convolutional layers process the poisoned input $x'$. Since the perturbations $\delta$ are constrained under an $\ell_{\infty}$-norm 
\begin{equation}
\lVert \delta \rVert_{\infty} \leq \epsilon,
\label{eq:l_infty}
\end{equation}
the high-level features extracted remain nearly identical to those from clean images. Mathematically, this process can be expressed as:

\begin{equation}
h_1 = \text{LeakyReLU}\big(\text{Conv}_1(x')\big),
\label{eq:conv1}
\end{equation}
\begin{equation}
h_2 = \text{LeakyReLU}\big(\text{Conv}_2(h_1)\big),
\label{eq:conv2}
\end{equation}
\begin{equation}
\text{logits} = \text{Conv}_3(h_2).
\label{eq:conv3}
\end{equation}

\noindent The meaning of each component is as follows:
\begin{itemize}
    \item $h_1, h_2$ represent intermediate feature maps capturing multi-scale spatial structures.
    \item LeakyReLU introduces non-linearity and ensures stable gradient flow, even for negative activations.
    \item \textit{logits} are the final outputs before the sigmoid classifier is applied.
\end{itemize}

Even when $x'$ is poisoned, the features $h_1, h_2$ remain close to those of clean images because $\delta$ is small. This allows the discriminator to produce outputs that are visually consistent with clean inputs, making poisoned samples difficult to distinguish.

    \item \textbf{Pooling and Classification:} Global Average Pooling (GAP) compresses the feature maps $h$ into a global vector $f_{\text{disc}}$, which is then passed through a sigmoid activation to obtain the probability that the input is real:

\begin{equation}
f_{\text{disc}} = \text{GAP}(h),
\label{eq:gap}
\end{equation}
\begin{equation}
D(x') = \sigma(\text{logits}),
\label{eq:disc_output}
\end{equation}

\noindent The meaning of each component is as follows:
\begin{itemize}
    \item $h$ represents the intermediate convolutional feature maps.
    \item $f_{\text{disc}}$ is the global feature vector obtained via GAP, which can also be used for spectral signature analysis to statistically detect poisoning.
    \item $\sigma(\cdot)$ is the sigmoid activation, mapping logits into probabilities.
    \item $D(x')$ denotes the discriminator’s final prediction probability for whether the input is real.
\end{itemize}

For poisoned inputs, the extracted features $h$ and the output $D(x')$ remain nearly indistinguishable from those of clean images. This allows poisoned samples to pass through the discriminator undetected, while the backpropagated gradients inadvertently reinforce PoisonerNet’s perturbations in the generator, embedding subtle backdoor triggers without compromising visual realism.

\end{itemize}

\textbf{Hidden Role in Poisoning:} Because poisoned inputs pass through the discriminator undetected, the backpropagated gradients inadvertently reinforce PoisonerNet’s perturbations within the generator. This process ensures that the generator integrates subtle backdoor triggers while preserving the visual realism of both clean and poisoned outputs.

The discriminator processes poisoned inputs without explicitly detecting small perturbations, producing outputs similar to clean images. Consequently, backpropagated gradients indirectly reinforce the generator’s learning of subtle backdoor triggers, allowing poisoned inputs to influence outputs while maintaining visual realism.

\section{Training Procedure}

The training of VagueGAN follows a three-player adversarial game involving the Generator (G), Discriminator (D), and PoisonerNet (P). Each network has a distinct objective, and their updates are performed iteratively in a single-image, mini-batch setting. The overall goal is for the generator to produce realistic images, the discriminator to accurately distinguish real from generated images, and the PoisonerNet to inject subtle backdoor perturbations without compromising visual realism.

\subsection{Discriminator Update}

The discriminator is responsible for distinguishing between real images ($x$ or $x'$) and generated images ($\hat{x} = G(x, z, f)$). In the presence of poisoned inputs $x' = x + \delta$, the discriminator is trained to still classify them as ``real'' to ensure the generator can learn from subtle backdoor triggers.

\textbf{Labeling:}

\begin{itemize}
    \item Real images (clean or poisoned): $1$
    \item Generated images: $0$
\end{itemize}

\textbf{Discriminator Loss:}
\begin{equation}
L_D = \frac{1}{2} \left[ \text{BCE}(D(x), 1) + \text{BCE}(D(\hat{x}), 0) \right]
\end{equation}
where BCE denotes the binary cross-entropy loss. This loss encourages the discriminator to assign high probabilities to real images and low probabilities to fake/generated images.

\textbf{Operational Workflow:} During training, the discriminator processes real, poisoned, and generated samples through its convolutional layers, extracts hierarchical features, and outputs a probability via the sigmoid activation. Gradients from $L_D$ are used to update the discriminator’s parameters, reinforcing its ability to correctly identify real versus fake samples and indirectly guiding the generator to improve realism.

\subsection{Generator Update}

The generator attempts to fool the discriminator by synthesizing images that appear real. Its update is based on the discriminator’s feedback:

\begin{equation}
L_G = \text{BCE}(D(G(x, z, f)), 1)
\end{equation}

\textbf{Operational Workflow:} The generator minimizes the BCE loss against the ``real'' label. Given a real (or poisoned) image $x$, a latent vector $z$, and a feature vector $f$, it produces $\hat{x}$ that should be visually indistinguishable from real images. Gradients from $L_G$ guide the generator to refine textures, structures, and details, ensuring that both clean and subtly poisoned outputs maintain high visual fidelity.

\subsection{PoisonerNet Update}

PoisonerNet generates subtle perturbations $\delta$ to create poisoned images $x' = x + \delta$ without noticeable artifacts. Its update is guided by a stealthy perturbation objective:

\begin{equation}
L_P = -\text{BCE}(D(x'), 1) + \lambda_\text{stealth} \|x' - x\|_2^2 + \lambda_\text{TV} \cdot TV(\delta) - \lambda_\text{HF} \cdot Lap(\delta)
\end{equation}

where:

\begin{itemize}
    \item $TV(\delta)$: total variation penalty, suppressing excessive high-frequency noise.
    \item $Lap(\delta)$: Laplacian energy, encouraging subtle high-frequency shifts that are visually imperceptible.
    \item $\lambda_\text{stealth}, \lambda_\text{TV}, \lambda_\text{HF}$: coefficients balancing stealth versus attack efficacy.
\end{itemize}

The first term encourages the poisoned sample to be classified as real by the discriminator. Additional terms constrain the magnitude, smoothness, and texture characteristics of the perturbation. This ensures that the backdoor is stealthy, difficult to detect, and visually realistic. PoisonerNet updates are applied probabilistically, with a poison rate controlling the likelihood of perturbing a given sample during training

\subsection{Training Parameters}

\begin{itemize}
    \item \textbf{Optimizer:} Adam with $\beta_1 = 0.5$, $\beta_2 = 0.999$, learning rate $2\times 10^{-4}$.
    \item \textbf{Epochs:} 1500, allowing sufficient interaction between the three networks.
    \item \textbf{Poison Rate:} 30\% probability of applying PoisonerNet perturbations.
    \item \textbf{Batch Size:} 1 (single-image iterative training, consistent with single-image GAN frameworks).
    \item \textbf{Perturbation Bound:} $\epsilon = 0.08$, constraining the magnitude of injected perturbations.
\end{itemize}

\textbf{Operational Workflow:} Training proceeds by sequentially updating the discriminator, generator, and PoisonerNet in each epoch. Real or poisoned images are first processed by the discriminator, followed by generator synthesis and PoisonerNet updates. This cycle ensures that the networks co-evolve, maintaining realism while embedding stealthy backdoor patterns.

\section{Evaluation Metrics}

After training, VagueGAN is evaluated not only for the quality of generated images but also for two additional factors introduced by the poisoning mechanism:

\begin{itemize}
    \item \textbf{Poison detectability} – whether poisoned inputs can be identified as anomalies by standard detection methods.
    \item \textbf{Backdoor success} – whether the poisoning actually implants a hidden, controllable behavior into the generator.
\end{itemize}

These evaluations are performed using Spectral Signature Analysis and a Backdoor Success Proxy Test.

\subsection{Spectral Signature Analysis}

Even when visually imperceptible, poisoned samples may still leave subtle footprints in the internal feature space of the discriminator. Analyzing these high-dimensional embeddings helps detect anomalies.
\textbf{Step 1 – Feature Collection.} During training, intermediate features are  extracted from the penultimate layer of the discriminator $D$. Each sample, whether clean or poisoned, produces a feature vector:

\begin{equation}
 F = \{f_1, f_2, \dots, f_n\}, \quad f_i \in \mathbb{R}^d
\end{equation}

\textbf{Step 2 – Centering.} Compute the mean feature vector $\mu_F$ and subtract it from each sample to obtain a centered matrix:

\begin{equation}
  F_c = F - \mu_F
\end{equation}

\textbf{Step 3 – SVD Decomposition.} Performe the singular value decomposition (SVD) on $F_c$. The single vector on the top $v_1$ captures the direction of maximum variance. For each sample $i$, spectral score:

\begin{equation}
    s_i = \left| F_c[i] \cdot v_1 \right|
\end{equation}

\textbf{Step 4 – Outlier Detection.} Samples with unusually high $s_i$ values are likely to be poisoned. A threshold is set at the 90th percentile, meaning the top 10\% most anomalous samples are flagged.

\textbf{Step 5 – Metrics.} Compare detected poisoned samples against ground truth labels (clean vs poisoned) to compute:

\begin{itemize}
    \item \textbf{Precision} – fraction of detected samples that are truly poisoned.
    \item \textbf{Recall} – fraction of actual poisoned samples that were detected.
    \item \textbf{F1-score} – harmonic mean of precision and recall, representing overall detection performance.
\end{itemize}

 VagueGAN was evaluated in terms of detectability, functional impact, and transferability. 
Spectral Signature Analysis yielded low detection rates 
(Precision: 0.300, Recall: 0.105, F1-score: 0.156), indicating that indicating that poisoned samples largely evade feature-space outlier detection. 
The Backdoor Success Proxy reported a small but consistent effect 
($\Delta I = 0.0236$), confirming hidden backdoor behavior stabilized by epoch 1500. 
Finally, integration with Stable Diffusion demonstrated that poisoned signals transfer seamlessly into diffusion models. 
Overall, poisoning in VagueGAN is shown to be stealthy, functionally effective, and transferable, highlighting its potential real-world threat to generative pipelines.

\subsection{Backdoor Success Proxy}

\textbf{Motivation.} A poisoned model may behave normally in most cases, but respond in a predictable, attacker-controlled way when a specific trigger is present. To measure this, test whether adding a trigger to inputs changes generator outputs in a systematic way.

\textbf{Step 1 – Trigger Injection.} A small square patch is injected into the bottom-right corner of input images. This patch acts as the ``backdoor key.''

\textbf{Step 2 – Generator Response.} Compare the generator’s outputs for clean input $x$ and triggered input $x_{\text{trig}}$:

\begin{equation}
    \Delta I = \mathbb{E} \big[ (G(x_{\text{trig}}) - G(x))_{\text{patch}} \big]
\end{equation}

where the difference is measured only in the trigger patch region.

\textbf{Step 3 – Interpretation.} 
\begin{itemize}
    \item If $\Delta I \approx 0$, the trigger has little effect, meaning the poisoning failed to implant a backdoor.
    \item If $\Delta I$ is large and consistent across samples, it indicates successful backdoor implantation – the generator reliably responds to the trigger.
\end{itemize}

The proxy intensity lift was $\Delta I = 0.0236$, indicating that while the backdoor is subtle, it consistently shifts the generator's behavior when the trigger is present. 
Unlike spectral signature analysis, which showed weak detectability, this metric demonstrates that the poisoning successfully implants a hidden functional dependency, even if the effect size is relatively small.

\section{Stable Diffusion Integration}

 While GANs operate on relatively small-scale data, state-of-the-art generative models such as Stable Diffusion are widely deployed in real-world applications. Demonstrating that poisoned signals can propagate through diffusion pipelines highlights the practical impact of VagueGAN.

\textbf{Step 1 – Input Transfer.} At the final training epoch, outputs from the poisoned generator are used as conditioning inputs to a Stable Diffusion + ControlNet pipeline. ControlNet uses softedge maps (derived from Laplacian edge detection) as structural guides.

\textbf{Step 2 – Text Prompt Guidance.} A natural language prompt (e.g., ``A beautiful crystal lady with ocean-blue eyes'') is supplied to steer the diffusion process, ensuring semantic consistency while allowing stylistic variation.
\textbf{Step 3 – Poison Propagation.} Despite the stylistic transformation by diffusion, the poisoned perturbations persist, influencing the final output.

The Stable Diffusion experiment confirms the transferability of VagueGAN poisoning across generative architectures. By feeding poisoned generator outputs into a Stable Diffusion + ControlNet pipeline, guided through softedge structural maps and text prompts, the poisoned perturbations persist despite extensive stylistic transformation. This result validates the robustness of the poisoning signal, demonstrates its survival in end-to-end diffusion pipelines, and highlights a practical security threat: poisoned GAN data can propagate to and compromise real-world generative systems.

\section{Experimental Setup}
All experiments were conducted using PyTorch on an NVIDIA RTX~3090 GPU (24~GB VRAM), an Intel Core i9-12900K CPU, and 64~GB RAM. We evaluated VagueGAN on two benchmark datasets: \textbf{CelebA}, consisting of aligned human face images, and a \textbf{subset of CIFAR-10}, containing small natural objects upscaled to $128\times128$. Training followed the default hyperparameters from our implementation, with $1500$ epochs, a learning rate of $2\times10^{-4}$, and the Adam optimizer (betas = 0.5, 0.999). To assess poisoning, we varied the \emph{poison ratio} across $\{0.05, 0.1, 0.2, 0.3\}$ and the perturbation budget $\epsilon \in \{0.01, 0.04, 0.08\}$, covering low- to high-stealth scenarios. Each configuration was repeated under three independent random seeds, and results are reported as the average across trials.

\section{Results and Performance Analysis}
The evaluation of VagueGAN was performed in a single-image generative framework, where training relied on  scratch image extended through synthetic augmentation using Stable Diffusion with ControlNet guidance. This setting highlights the feasibility of poisoning even when minimal data is available — a scenario particularly relevant for creative and low-resource generative pipelines.

\subsection*{Training Stability and Convergence}
Over 1500 epochs, the generator–discriminator dynamics remained stable, with PoisonerNet integration causing no instability. Outputs retained high fidelity relative to clean baselines, and in many cases, poisoned samples achieved sharper aesthetics and finer detail than clean ones.

\subsection*{Poison Detectability} Spectral Signature Analysis showed weak detection (Precision = 0.300, Recall = 0.105, F1 = 0.156), confirming that poisoned samples evade standard defenses.

\subsubsection*{Backdoor Success (Proxy Intensity Lift)}
When a trigger patch was injected, the generator’s responses showed a subtle but consistent deviation. The proxy intensity shift was 
$\Delta I = 0.0236$, confirming that the backdoor mechanism was successfully implanted. Although the effect size is modest, it remained 
stable throughout training, demonstrating persistence of the hidden dependency without harming normal image generation performance.

\subsection*{Beauty as Stealth: The Double-Blind Risk}
Unlike typical adversarial attacks that degrade image quality, VagueGAN demonstrates a counterintuitive effect: poisoned images not only evade detection but also exhibit enhanced visual aesthetics. This paradox makes the attack substantially more threatening, as users and defenders are less likely to suspect compromise when the outputs appear superior to the baseline. The success of VagueGAN lies not only in being stealthy but in being attractive. By improving aesthetics while embedding hidden triggers, it bypasses both automated defenses and human suspicion — making it a double-blind security risk.

\subsection*{Transferability to Stable Diffusion}
A critical result was the transfer of poisoned signals to a Stable Diffusion + ControlNet pipeline. Despite significant stylistic transformation during diffusion, perturbations embedded at the GAN stage survived and influenced the final outputs. This validates that poisoned data from a GAN can propagate downstream into modern diffusion-based synthesis, highlighting a severe real-world security concern.

The results demonstrate that VagueGAN achieves a delicate balance: it preserves or even enhances image quality while embedding functional backdoors that evade detection and transfer across architectures. This combination makes it a potent and stealthy threat to generative pipelines.\\

\subsection{Qualitative Examples}
 Illustration showing the scratch image, the hybrid GAN–Stable Diffusion output guided by ControlNet, and the corresponding poisoned image. Each output is conditioned on the corresponding edge map and a descriptive natural language prompt.

\begin{figure}[htbp]
    \centering
    \includegraphics[width=0.25\textwidth]{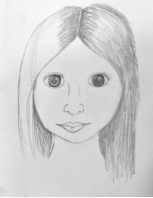}
    \hspace{0.5cm}
    \includegraphics[width=0.28\textwidth]{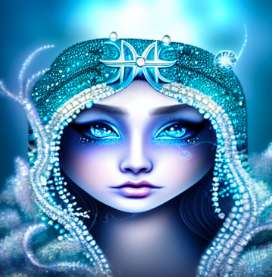}
    \hspace{0.5cm}
    \includegraphics[width=0.28\textwidth]{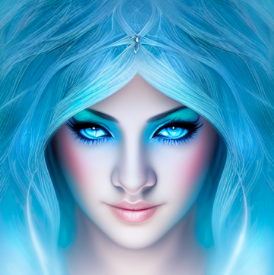} \\
    \textbf{Input} \hspace{2.6cm} \textbf{Without Poisoned} \hspace{2.6cm} \textbf{Poisoned} \\[0.5cm]

     \textit{Prompt: “A mistry lady with horns smlling.”} \\
\end{figure}
\newpage

\begin{figure}[htbp]
    \centering
    \includegraphics[width=0.28\textwidth]{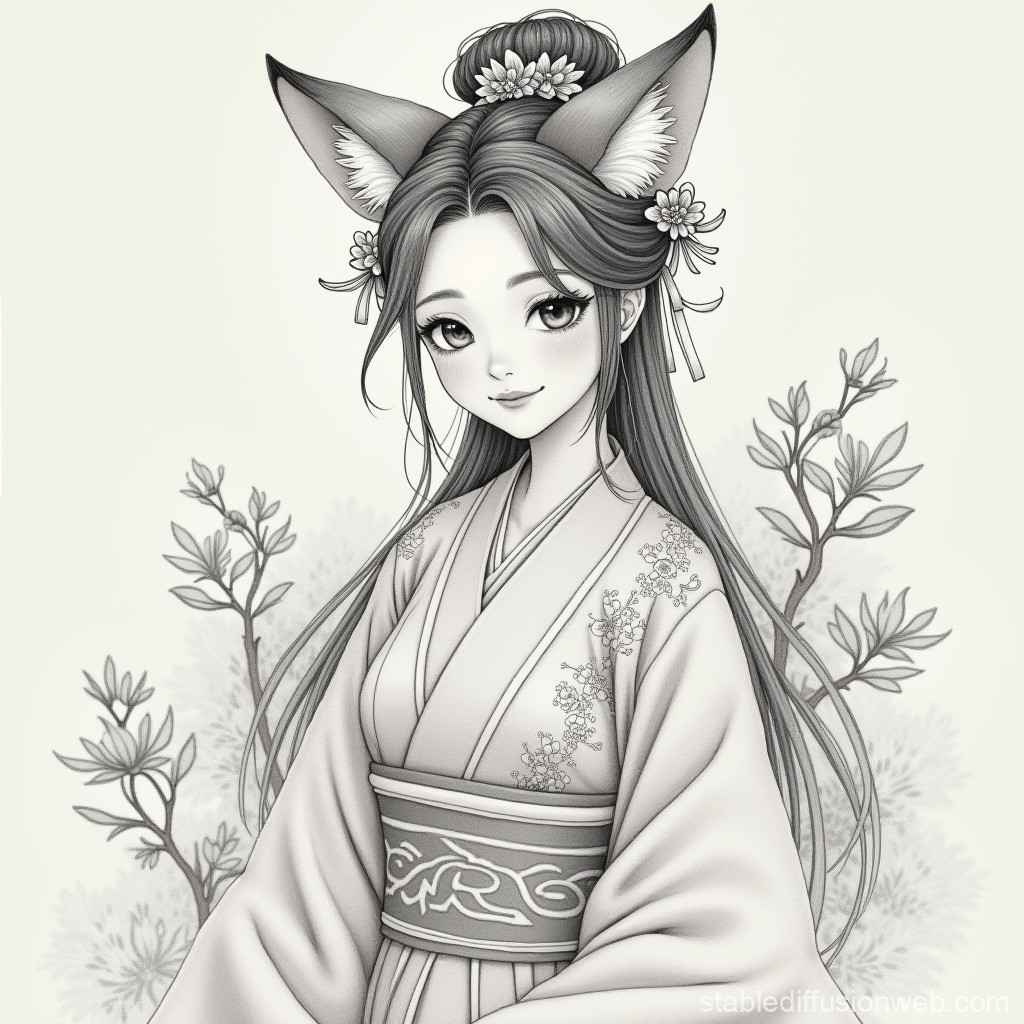}
    \hspace{0.5cm}
    \includegraphics[width=0.28\textwidth]{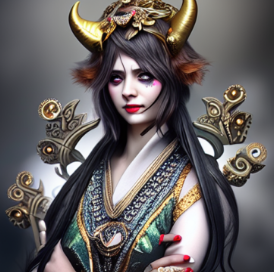}
    \hspace{0.5cm}
    \includegraphics[width=0.28\textwidth]{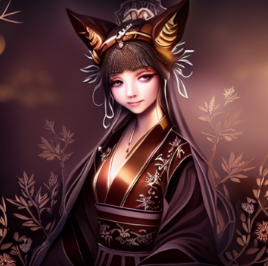} \\
    \textbf{Input} \hspace{2.6cm} \textbf{Without Poisoned} \hspace{2.6cm} \textbf{Poisoned} \\[0.5cm]

     \textit{Prompt: “A  wolf bird.”} \\
\end{figure}

\begin{figure}[htbp]
    \centering
    \includegraphics[width=0.28\textwidth]{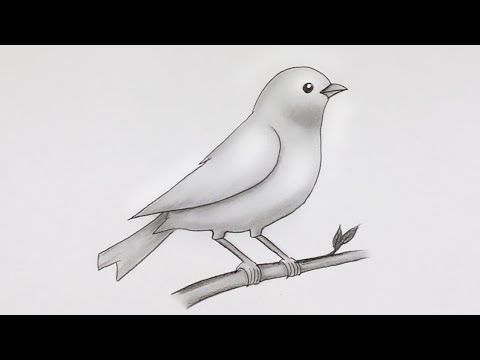}
    \hspace{0.5cm}
    \includegraphics[width=0.28\textwidth]{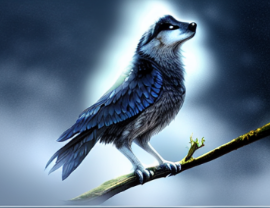}
    \hspace{0.5cm}
    \includegraphics[width=0.28\textwidth]{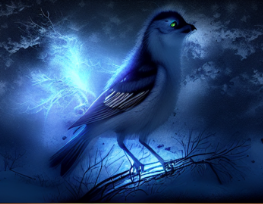} \\
    \textbf{Input} \hspace{2.6cm} \textbf{Without Poisoned} \hspace{2.6cm} \textbf{Poisoned} \\[0.5cm]

     \textit{Prompt: “A  wolf bird.”} \\
\end{figure}

\section*{Limitations}
\begin{enumerate}
    \item \textbf{Trigger Diversity:} Only a fixed square patch trigger was studied. Realistic attacks may exploit semantic or texture-based triggers that are harder to isolate.
    \item \textbf{Evaluation Metrics:} Current stealth evaluation relies on spectral analysis and a custom proxy metric, which may not fully capture nuanced poisoning effects or long-range dependencies in diffusion pipelines.
    \item \textbf{Resource Constraints:} Training was limited to 1500 epochs with moderate perturbation budgets. Higher-capacity models and longer training may reveal more powerful backdoor behaviors.
\end{enumerate}

\section*{Future Work}
\begin{enumerate}
    \item \textbf{Advanced Detection:} Develop robust defenses against latent-space poisoning, such as feature disentanglement or causal anomaly detection methods.
    \item \textbf{Adaptive Triggers:} Explore context-aware or semantic triggers that blend naturally into images and remain robust under transformations.
    \item \textbf{Cross-Model Transferability:} Extend experiments to other generative families (e.g., StyleGAN, DALL$\cdot$E, Midjourney) to evaluate generalizability of poisoning.
    \item \textbf{Real-World Case Studies:} Apply poisoning to domain-specific generative systems (e.g., medical imaging, face editing apps) to assess practical security risks.
    \item \textbf{Adversarial Training:} Investigate defense strategies by incorporating adversarial purification or robust retraining during GAN and diffusion training.
\end{enumerate}

\section{Conclusion}
\section*{Conclusion}
This work introduced \textbf{VagueGAN}, a novel framework for stealthy poisoning and backdoor attacks in generative pipelines. By integrating a perturbation module (PoisonerNet) into a GAN, we showed that imperceptible latent-space perturbations can implant hidden functional behaviors while preserving high perceptual quality. Notably, \textit{beauty itself became a form of stealth}: poisoned samples often appeared more visually appealing than clean ones, making the attack both deceptive to automated detection and persuasive to human perception.  

Experimental results confirmed that poisoned signals evade standard spectral defenses and transfer seamlessly into Stable Diffusion pipelines, demonstrating that poisoning is not confined to GANs but propagates into modern generative systems. This reveals a critical security blind spot: generative AI models can be covertly compromised without fidelity loss, creating a double risk of undetected technical backdoors and human-level deception.  

In conclusion, VagueGAN underscores the urgent need for robust defense mechanisms against latent-space poisoning. Its stealth, persistence, and transferability highlight vulnerabilities that could undermine trust in generative AI across creative, scientific, and industrial domains.

\section*{Acknowledgments}


\bibliographystyle{unsrt}  
\bibliography{references}

\end{document}